\documentclass[letterpaper, 10 pt, journal, twoside]{IEEEtran}
\IEEEoverridecommandlockouts                              % This command is only
\usepackage{xcolor}
\usepackage{comment}
\usepackage{graphics} % for pdf, bitmapped graphics files
\usepackage{epsfig} % for postscript graphics files
\usepackage{amsmath} % assumes amsmath package installed
\usepackage{amssymb}  % assumes amsmath package installed
\usepackage{mathtools}
\usepackage{marginnote}
\usepackage{hyperref}
\usepackage{cite}
\usepackage{mathrsfs}
\usepackage{algorithm,algorithmic}
\usepackage{xcolor}
\usepackage[inkscapelatex=false]{svg}
\usepackage{multirow} % Add this to the preamble
\usepackage{array} 
\usepackage{etoolbox}

\makeatletter
\patchcmd{\@makecaption}
  {\scshape}
  {}
  {}
  {}
\makeatletter
\patchcmd{\@makecaption}
  {\\}
  {.\ }
  {}
  {}
\makeatother
\def\tablename{Table}

\title{\LARGE \bf
Fast Decentralized State Estimation for Legged Robot Locomotion via EKF and MHE} % may change the name 

\author{ Jiarong Kang$^{*}$, Yi Wang$^{*}$, and Xiaobin Xiong   % stops a space
\thanks{Manuscript received: May, 29, 2024; Revised: August, 25, 2024; Accepted: September, 28, 2024.}
\thanks{This paper was recommended for publication by Abderrahmane Kheddar
 upon evaluation of the Associate Editor and Reviewers' comments.}
\thanks{$^*$ The authors contribute equally to this work.}
\thanks{The authors are with the Wisconsin Expeditious Legged Locomotion (WELL-Lab) at the University of Wisconsin-Madison.
        {Corresponding to \tt\small xiaobin.xiong@wisc.edu}.
The experiment video can be seen here: \href{https://youtu.be/k3bBO87UIlk}{\texttt{https://youtu.be/k3bBO87UIlk}}
}%
}

\begin{document}

\newcommand{\Kang}[1]{{\color{blue} #1}}
\newcommand{\block}[1]{\noindent{\textbf{#1}:}}
\newcommand{\emphhh}[1]{{\color{yellow} \textbf{#1}}}

\markboth{IEEE Robotics and Automation Letters. Preprint Version. Accepted September, 2024}
{Kang \MakeLowercase{\textit{et al.}}: Fast Decentralized State Estimation for Legged Robot Locomotion via EKF and MHE} 

\maketitle
% \thispagestyle{empty}
% \pagestyle{empty}

%%%%%%%%%%%%%%%%%%%%%%%%%%%%%%%%%%%%%%%%%%%%%%%%%%%%%%%%%%%%%%%%%%%%%%%%%%%%%%%%
\begin{abstract}

In this paper, we present a fast and decentralized state estimation framework for the control of legged locomotion. The nonlinear estimation of the floating base states is decentralized to \textit{an orientation estimation via Extended Kalman Filter (EKF)} and \textit{a linear velocity estimation via Moving Horizon Estimation (MHE)}. The EKF fuses the inertia sensor with vision to estimate the floating base orientation. The MHE uses the estimated orientation with all the sensors within a time window in the past to estimate the linear velocities based on a time-varying linear dynamics formulation of the interested states with state constraints. More importantly, a marginalization method based on the optimization structure of the full information filter (FIF) is proposed to convert the equality-constrained FIF to an equivalent MHE. This decoupling of state estimation promotes the desired balance of computation efficiency, accuracy of estimation, and the inclusion of state constraints. The proposed method is shown to be capable of providing accurate state estimation to several legged robots, including the highly dynamic hopping robot PogoX, the bipedal robot Cassie, and the quadrupedal robot Unitree Go1, with a frequency at 200 Hz and a window interval of 0.1s. 
\end{abstract}

\section{Introduction}

Legged robots have undergone comprehensive development in several aspects, including design, manufacturing, control, and perception. They have shown great capabilities and future potentials in addressing societal problems such as these in logistics, search and rescue, inspection, and entertainment\cite{7758092}. With the demand for robot applications in outdoor environments, fast state estimation leveraging onboard sensors is crucial for realizing autonomous behaviors.

% for the success of the deployment. 

% Generally, the robot state denotes the base pose and velocity. 
Due to the inherently underactuated nature of legged robots, an accurate and high-frequency estimation of the torso velocity and orientation is essential for the control of dynamic locomotion \cite{bloesch2013state}, \cite{Wisth_2020}. This is a ubiquitous requirement for all dynamic legged robots regardless of how many legs they have. Extended Kalman filter (EKF) and its variants \cite{bloesch2013state,hartley2019contactaided,6942674} are commonly used to solve this problem. These methods provide an accurate recursive solution for the fusion of onboard proprioceptive sensors including Inertial Measurement Units (IMU) and joint encoders. However, when working with locomotion behaviors associated with longer aerial phases, measurements from leg kinematics cannot be used to correct the floating base states. With only inertial sensors, the estimation of velocities can quickly diverge. Additionally, the yaw drift is inevitable when only using proprioceptive sensors \cite{bloesch2013state}. To address this problem, several works \cite{10.3389/frobt.2020.00068,teng2021legged} have included exteroceptive sensors such as cameras and Lidar within the filtering methods. However, \cite{10.3389/frobt.2020.00068} primarily focused on pose drift correction, and \cite{teng2021legged} integrated vision and IMU into a single measurement, which compromised the robustness of the estimation due to potential correlations. %%% last sentence is unclear, and missing citation

%%% the starting sentence requires work. It's not a rigorous claim. 
To address this limitation, smoothing method that utilizes windowed optimization to include a history of sensor measurements is advantageous since multirate sensor data can be fused in the window. For instance, \cite{Wisth_2020} utilizes a factor graph method, which tightly couples vision with IMU and encoder measurements through preintegration. However, in robotic implementations, factor graph-based methods typically have high computational demands, making their estimation rate insufficient for providing optimization results for real-time control. Separate EKFs are often employed in between the factor graph optimization to provide high-frequency estimates \cite{Wisth_2020}. Another smoothing method, Moving Horizon Estimation (MHE), known as the dual problem of Model Predictive Control (MPC), provides an alternative formulation of the windowed optimization. Similar to MPC \cite{8594448}, when the dynamics is nonlinear, the MHE becomes a nonconvex optimization which yields demanding computation. Inspired by MHE, \cite{6942679} proposed a Quadratic Program (QP) based state estimator using full-body dynamics with state constraints, to estimate contact force and joint torques. An inequality-constrained MHE was explicitly applied in \cite{BAE20173793} for a bipedal robot, demonstrating its ability to handle constraints and non-Gaussian noises. However, only proprioceptive sensors were considered in \cite{BAE20173793}, and measurements prior to the smoothing window were directly discarded, which compromised the optimality of the estimation.

\begin{figure}
    \centering
    % \setlength{\abovecaptionskip}{0pt}
    % \includesvg[width=0.95\linewidth]{figure/first_block.svg}
    \includegraphics[width=0.95\linewidth]{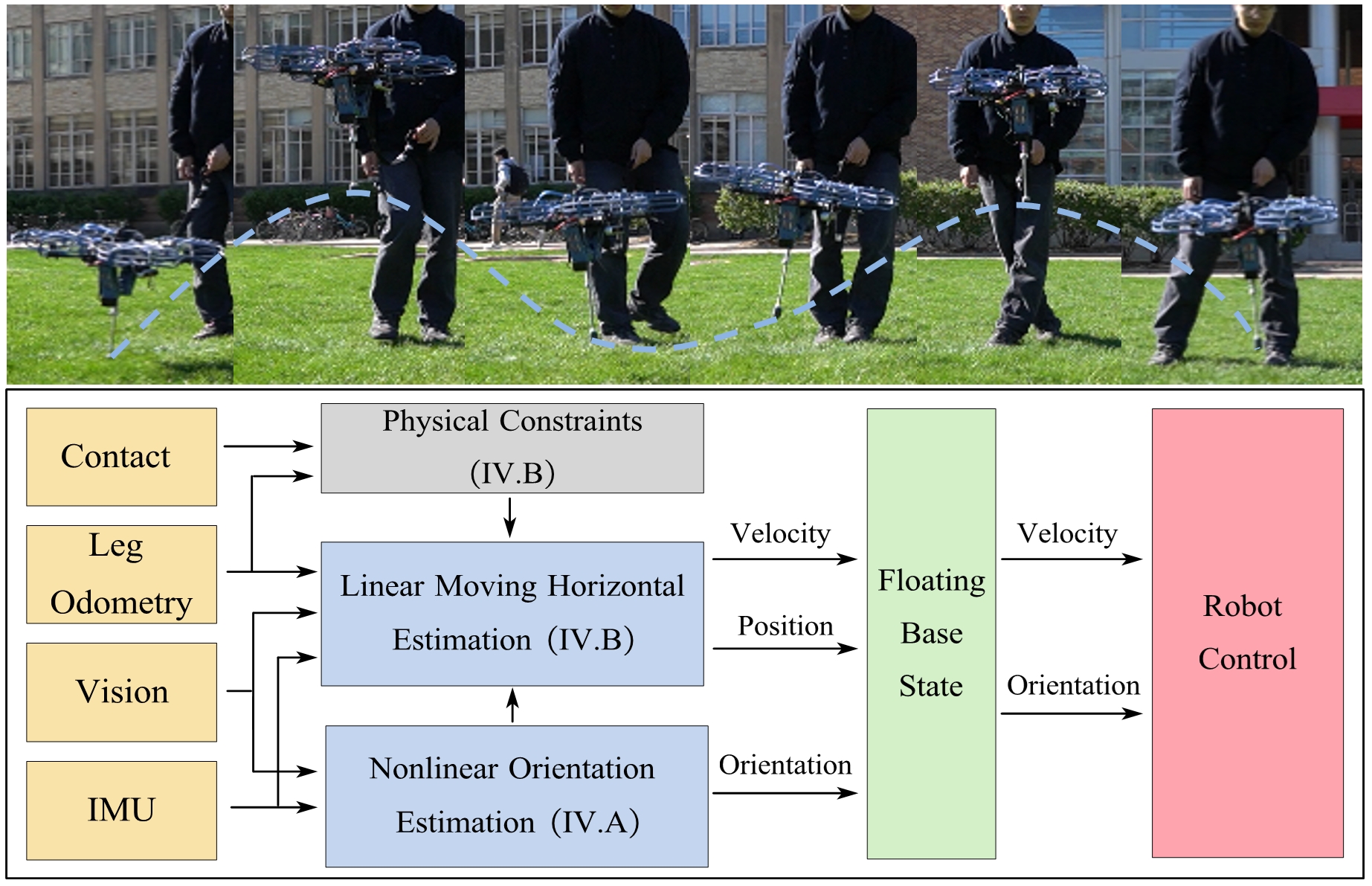}
    \vspace{-5pt}
    \caption{The proposed estimation framework (bottom) and its application for realizing dynamic hopping on PogoX (top).}
    \label{PogoX}
    \vspace{-20pt}
\end{figure}

% To the best of our knowledge, no prior work has systematically introduced the MHE framework to the field of legged robots. 
In this paper, we aim to develop a MHE based framework that achieves multirate sensor fusion of both proprioceptive and exteroceptive sensors, providing fast and accurate estimation for legged robot control. The computation burden of nonlinear optimization of the canonical MHE is addressed by the decentralization of the orientation and velocity estimation: the nonlinear orientation estimation relies on an EKF, whereas the linear velocity estimation is solved via the MHE. The framework is illustrated in Fig. 1, and it is applicable to legged robots that have IMUs, leg joint encoders, and cameras. The work has the following contributions:

\begin{itemize}
    \item A novel state estimation framework, integrating a lightweight orientation EKF with a linear MHE for velocity estimation, to seamlessly fuse multirate proprioceptive and exteroceptive sensors while incorporating state constraints in real-time.
    \item An extension of the marginalization method from \cite{5980267} to the equality-constrained MHE problem, resulting in a novel arrival cost calculation for the MHE formulation under equality constraints.
    \item Evaluations of the proposed decentralized estimators on various legged robotics systems, including the unipedal hopping robot PogoX \cite{wang2023terrestrial}, bipedal robot Cassie \cite{hartley2019contactaided}, and quadrupedal robot Unitree Go1 \cite{UNTREEWebsite}, with software implementation open-sourced at \cite{Github}.
\end{itemize}

\section{Related Work}

State-of-the-art methods for legged robot state estimation are categorized into filtering and smoothing approaches. For filtering approaches, they typically perform fusions of proprioceptive sensors (IMUs, force/torque sensors, and joint encoders) with high frequency (100 - 2000 Hz) using Kalman Filters (KFs). \cite{bloesch2013state} showed that using EKF, the robot pose, velocity, and IMU bias could be estimated on quadrupedal robots. An Invariant-EKF \cite{hartley2019contactaided} was proposed to improve the convergence of the estimation. The filtering approaches have also been widely implemented on bipedal robots \cite{6942674}. While providing high-frequency velocity estimation, the estimation accuracy is strongly affected by the availability of foot contact and accurate leg kinematics. For robots that perform highly dynamic locomotion with long aerial phases, the filtering approaches using only proprioceptive sensors can result in incorrect or biased velocity estimation \cite{teng2021legged}. 

Recent research on legged state estimation via smoothing utilizes the techniques from Simultaneous Localization And Mapping (SLAM) and Micro Aerial Vehicle (MAV) communities. The factor graph formulation, or specifically the Visual-Inertial Odometry (VIO) \cite{9196524}, incorporating the exteroceptive inputs, has successfully been applied to the control of aerial vehicles. In the field of legged robotics, the inclusion of leg kinematics yields the Visual-Inertial-Leg-Odometry (VILO) estimator.
% \cite{Wisth_2019} utilized a Two-State Implicit Filter (TSIF) \cite{Bloesch2018TheTI} to fuse the Leg Odometry (LO) and deployed the VILO at an industrial facility. 
\cite{Wisth_2020} and \cite{Hartley_2018} developed a contact preintegration algorithm, 
achieving accurate pose estimation with low drift. However, the formulation of the factor graph restricts the acquisition of fast velocity estimates, since the optimization results can only occur at image frames (10 - 60 Hz). Generally speaking, as shown in \cite{Wisth_2020} and \cite{Hartley_2018}, filter-based estimators are suitable to forward propagate high-frequency estimates. Given their recursive formulation, incorporating state constraints and managing multirate sensors can be cumbersome. Our approach aims to achieve high-frequency estimation through a smoothing method, fully leveraging the advantages of windowed optimization to incorporate multirate sensor readings and state constraints.

\section{Preliminaries}
We first present the preliminaries of Moving Horizon Estimation \cite{1178905} and the process and measurement models for legged robot state estimation \cite{bloesch2013state,10.3389/frobt.2020.00068,teng2021legged}.
\subsection{Moving Horizon Estimation}

A general description of the dynamic system involves the nonlinear dynamics and measurement equation:
\begin{align}
         x^{+} = f(x,u) + w,  \quad y = h(x) + v , \label{eq:nonlineardiff} 
\end{align}
where $(\cdot)^+$ denotes the updated state after a fixed interval $\Delta t$, $f(\cdot)$ and $h(\cdot)$ denote the process model and measurement model, respectively. $x$, $u$ and $y$ denote the state, control input and measurement of the system, respectively. $w$ and $v$ represent the process noise and measurement noise. 

Given a history of measurements $y_{[0,T]} = [ y_0^\intercal \  y_1^\intercal \  ... \  y_{T}^\intercal ]^\intercal$, the state estimation problem is to estimate the system state at this time instant $T$, denoted by $x_T$, or the state trajectory starting from the beginning up to this instant $T$, denoted by $x_{[0,T]} = [ x_0^\intercal \  x_1^\intercal \  ... \  x_{T}^\intercal ]^\intercal$. The notation $(\cdot)_{[i,j]}$ denotes the sequence of vectors that starts from time index $i$ and ends at time index $j$. The optimal state estimator solves the Maximum A Posterior (MAP) problem:
\begin{equation}
    x_{[0,T]}^*
   =  \underset{x_{[0,T]}}{\text{argmax}}\ {p(x_{[0,T]} | y_{[0,T]})} \label{eq:MAP} ,
\end{equation}
where $p(x_{[0,T]} | y_{[0,T]})$ is the posterior probability of the state trajectory $x_{[0,T]}$, and $(\cdot)^*$ denotes the optimal solution. When system is observable, the MAP problem \eqref{eq:MAP} can be solved via the FIF formulation \cite{FIF}, with model descriptions in \eqref{eq:nonlineardiff} being formulated as state constraints: 
\begin{align}
\min_{\mathcal{X}_{[0,T]}}\  \Gamma(x_0) &+ \sum_{k = 0}^{T-1}{w_k^\intercal Q_k^{-1} w_k} + \sum_{k=0}^T{v_k^\intercal R_k^{-1} v_k} ,  \tag{FIF} \label{eq:FIF}\\
\text{s.t.} \quad x_{k+1} &= f(x_k, u_k) + w_k,\ \forall k \in \{0,...,T-1 \}, \label{eq:dynamic_constraints}\\
 y_k &= h(x_k) + v_k,\ \forall k \in \{0,...,T \},\label{eq:observation_constraints} \\
 g(x_k) &\leq 0, \ \forall k \in \{0,...,T \}, \label{eq:inequality_constraints}
\end{align}
where $\mathcal{X}_{[0,T]}$ is the state and noise trajectories up to $T$:

\begin{align*}
    \mathcal{X}_{[0,T]} &= \begin{bmatrix}
    x_{[0,T]}^\intercal\ 
    w_{[0,T-1]}^\intercal\    
    v_{[0,T]}^\intercal
    \end{bmatrix}^\intercal, \\
    w_{[0,T-1]} &= \begin{bmatrix}
        w_0^\intercal &  w_1^\intercal & ... & w_{T-1}^\intercal
    \end{bmatrix}^\intercal, \\
    v_{[0,T]} &= \begin{bmatrix}
        v_0^\intercal &  v_1^\intercal & ... & v_{T}^\intercal
    \end{bmatrix}^\intercal,
\end{align*}
% \end{small}%
$g(\cdot)$ denotes general state constraints, and $Q$ and $R$ denote the covariance matrices of the process and measurement noises, respectively. Given a Gaussian-distributed prior $x_0 \sim \mathcal{N}(x_{\text{prior}}, P_0)$, the prior cost $\Gamma(x_0)$ is calculated as:
\begin{equation}
    \Gamma(x_0) = (x_0 - x_{\text{prior}})^\intercal P_0^{-1} (x_0 - x_{\text{prior}}) \label{eq:arrival0}. 
\end{equation}
FIF does not require the assumption of normally distributed noise; yet, we follow the Gaussian-distributed process and measurement noises assumption from \cite{bloesch2013state,10.3389/frobt.2020.00068,teng2021legged}.

The major drawback of FIF is that the computation cost increases w.r.t. time $T$, therefore, making it computationally intractable. To address this, Moving Horizon Estimation (MHE) is developed to use a fixed window of data in the past, thereby yielding a bounded computation cost. It is formulated as repeatedly solving this optimization problem: 
\begin{figure}
    \centering
    \setlength{\abovecaptionskip}{0pt}
    % \includesvg[width=0.95\linewidth]{figure/MHE_explain.svg}
    \includegraphics[width=0.95\linewidth]{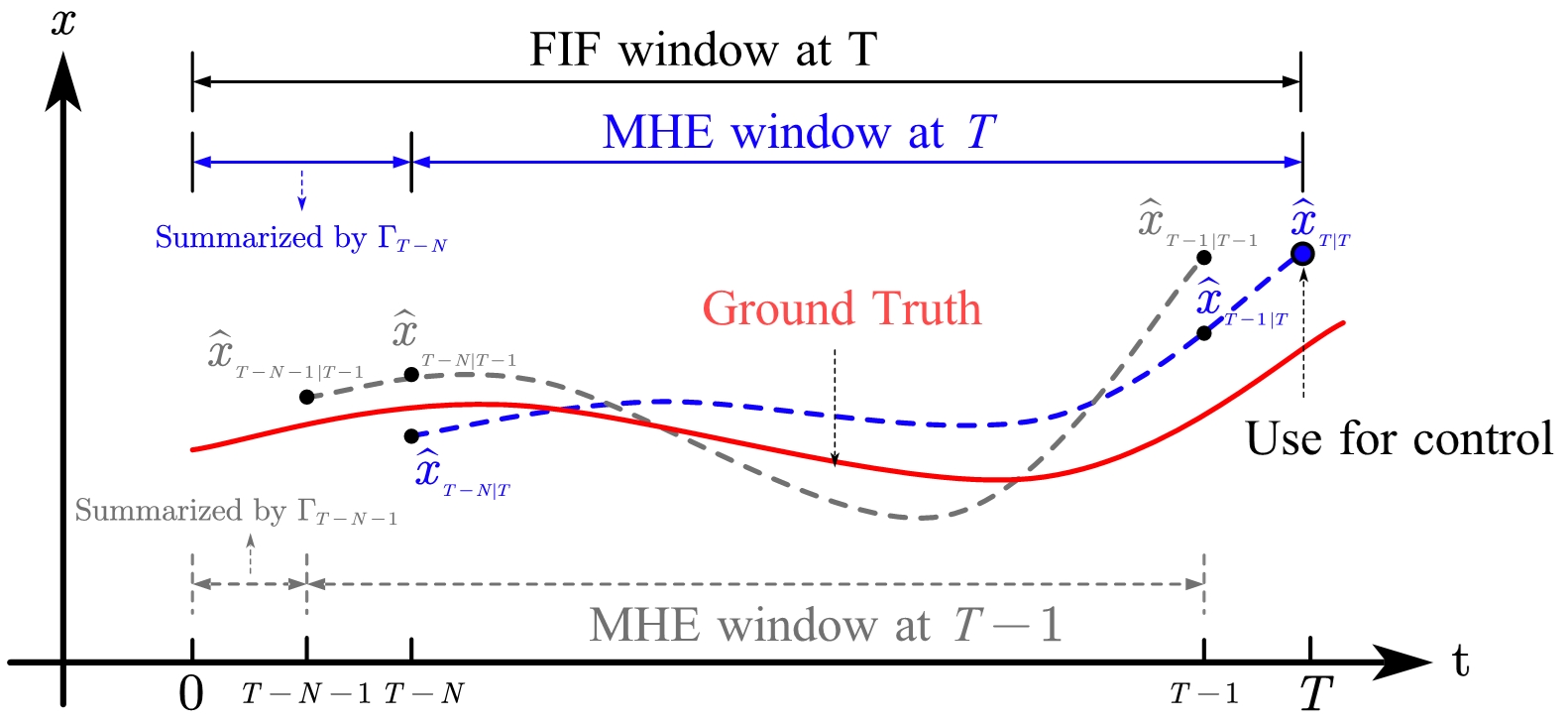}
    \vspace{-2pt}
    \caption{Illustration of MHE, arrival cost $\Gamma$ and their relationship with FIF, where $\hat{x}_{i|k}$, $\{i,k \in \mathbb{N}^+|\ 0 \leq i \leq k \leq T \}$ is the estimate of $x$ at time index $i$, using measurements from time index 0 to time index $k$.}
    \vspace{-15pt}
    \label{MHE_illu}
\end{figure}
\begin{align}
 {\hat{\mathcal{X}}_{[T-N,T]}} =\  &  \underset{\mathcal{X}_{[T-N,T]}}{\text{argmin}} \  \Gamma(x_{T-N}) +  \textstyle{\sum}_{k = T-N}^{T-1}{w_k^\intercal Q_k^{-1} w_k} \textstyle \nonumber \\ 
& + \textstyle {\sum}_{k=T-N}^T{v_k^\intercal R_k^{-1} v_k}  ,  \tag{MHE} \label{eq:MHE} \\ 
\text{s.t.} &\quad \eqref{eq:dynamic_constraints} ,\  \eqref{eq:observation_constraints} ,\ \eqref{eq:inequality_constraints},\ k \geq T-N , \nonumber  %%% I don't think this is a good notation
\end{align}
where $\hat{(\cdot)}$ denotes the estimated variables, and $\mathcal{X}_{[T-N,T]}$ contains the state and noises trajectories starting from time index $T-N$ up to time index $T$:
\begin{equation*}
\textstyle
    \mathcal{X}_{[T-N,T]} = \begin{bmatrix}
    x_{[T-N,T]}^\intercal\ 
    w_{[T-N,T-1]}^\intercal\    
    v_{[T-N,T]}^\intercal
    \end{bmatrix}^\intercal.
\end{equation*}
The basic concept of the MHE is illustrated in Fig. \ref{MHE_illu}: the arrival cost $\Gamma(x_{T-N})$ serves as an approximation metric to FIF \cite{1178905}, which imposes previous information prior to the current window to the mean and covariance of $x_{T-N}$. The arrival cost is calculated as described in \eqref{eq:arrival0}, where the prior mean and covariance are typically calculated via EKF \cite{1178905}. 

From an algorithmic perspective, MHE is often considered as the dual of MPC, solving which generally requires a trade-off between modeling accuracy and computational efficiency. Similarly, despite having a bounded computational cost, nonlinear MHE problems have challenges in keeping up with the sampling frequencies of proprioceptive sensors.

\subsection{Legged Robotic Estimation Model}
We now review the common practices of state estimation for legged robots. A set of robot-centric states is chosen to describe the motion of the legged robot, as in \cite{bloesch2013state,10.3389/frobt.2020.00068, teng2021legged}:
\begin{equation}
    \mathbf{x} := \begin{bmatrix}
        p^{\intercal} & v^{\intercal} & q^{\intercal} & p_{\text{foot}}^{\intercal} & b_a^{\intercal} & b_{\omega}^{\intercal} 
    \end{bmatrix}^{\intercal}, \label{eq:state}
\end{equation}
where $p$, $v$, and $p_{\text{foot}}$ denote the base position, base velocity, and foot position, respectively, in the world frame denoted by $W$. $q \in \mathbb{H}$ denotes the quaternion representation of the orientation of the base $ R_{\text{WB}} \in SO(3)$,
and $b_a$ and $b_{\omega}$ represent the accelerometer bias and gyroscope bias, respectively, in the body frame $B$. Only one of $N$ legs is included here for notation simplicity. The actual number of feet depends on the robot during its locomotion behaviors.
\subsubsection{Sensors Models} The following sections describe the models of the sensors used on the robot.
\newline
\noindent{\textbf{Inertial Sensors:}}
The IMU measures the acceleration $a$ and angular rate $\omega$ of the floating base in the body frame $B$. 
The following model is widely used to describe this process:
\begin{equation}
    \tilde{a} = a + b_a + \delta_a, \quad \tilde{\omega} = \omega + b_{\omega} + \delta_{\omega},    
\end{equation}
in which the measured inertial readings $\tilde{a}$ and $\tilde{\omega}$ are corrupted by Gaussian noises $\delta_{a} \sim \mathcal{N}(0,Q_{a})$ and $\delta_{\omega} \sim \mathcal{N}(0,Q_{\omega})$, respectively. The additive bias terms $b_{a}$ and $b_{\omega}$ follow the random walk model, and they have discrete dynamics as:
\begin{equation}
    b_a^+ = b_a + \delta_{b_a}, \quad b_\omega^+ = b_{\omega} + \delta_{b_{\omega}},
\end{equation}
where $\delta_{b_a} \sim \mathcal{N}(0, Q_{b_a}) $ and $\delta_{b_{\omega}} \sim \mathcal{N}(0, Q_{b_{\omega}})$ are the additive white Gaussian noises, respectively.

\noindent{\textbf{Joint Encoders and Leg Kinematics:}}
The legged robot is equipped with joint encoders to provide information of the joint angles $\alpha$ and joint angular velocities $\dot{\alpha}$. The measurements are assumed to be corrupted by discrete Gaussian noises $\delta_{\alpha} \sim \mathcal{N}(0,Q_{\alpha})$ and $\delta_{\dot{\alpha}} \sim \mathcal{N}(0,Q_{\dot{\alpha}})$, respectively:
\begin{equation}
        \tilde{\alpha} = \alpha + \delta_{\alpha}, \quad \dot{ \tilde{\alpha}} = \dot{\alpha} + \delta_{\dot{\alpha}}. 
\end{equation}
Based on the known leg kinematics, the relative position $p_{\text{foot}} - p$ and relative velocity $v_{\text{foot}} - v$ of the foot w.r.t. the base, in the body frame $B$, are computed as:
\begin{align}
    R_{\text{WB}}^T (p_{\text{foot}}-p) &=  fk(\tilde{\alpha}) + \delta_{\text{pf}}, \label{eq:pfoot} \\
    R_{\text{WB}}^T(v_{\text{foot}}-v) &=  J(\tilde{\alpha}) \tilde{\dot{\alpha}} + (\tilde{\omega} - b_{\omega})^{\times}  fk(\tilde{\alpha}) + \delta_{\text{vf}}, \label{eq:vfoot}
\end{align}
where $fk(\cdot)$, $J(\cdot)$ denote the forward kinematics and Jacobian of the corresponding foot. $(\cdot)^{\times}$ maps $\omega$  to the corresponding skew-symmetric matrix.  $\delta_{\text{pf}} \sim \mathcal{N}(0,Q_{\text{pf}})$ and $\delta_{\text{vf}} \sim \mathcal{N}(0,Q_{\text{vf}})$ denote the combinations of multiple uncertainties \cite{teng2021legged}, including the calibration and kinematics modeling error, encoder noises and gyroscope noises. 
% $\delta_{pfk} \sim \mathcal{N}(0,Q_{pfk})$, $\delta_{vfk} \sim \mathcal{N}(0,Q_{vfk})$, 

% If we assume the contact point remains static with respect to the world frame $W$, static contact constraints are used to formulate the Leg Odometry (LO) model.
\subsubsection{Process Model}
The state evolution of \eqref{eq:state} is modeled by the following nonlinear discrete process model:
\begin{equation}
    \mathbf{x}^+ = \begin{bmatrix}
        p + v \Delta t + \frac{1}{2} \big(R_{\text{WB}} (\tilde{a} - b_{a}) + \text{g} \big) \Delta t^2\\
        v + \big(R_{\text{WB}} (\tilde{a} - b_{a}) + \text{g}\big) \Delta t\\
        \zeta \big((\tilde{\omega}-b_{\omega}) \Delta t \big)\otimes q\\
        \begin{bmatrix}
        p_{\text{foot}}^\intercal & b_{a}^\intercal &b_{\omega} ^\intercal
        \end{bmatrix}^\intercal
    \end{bmatrix} + \delta_{\mathbf{x}} ,\label{eq:nonlinear_dyn}
\end{equation}
where $\otimes$ denotes the quaternion multiplication, $\zeta(\cdot)$ is the map from the error rotation to the error quaternion:
\begin{align}
    \zeta(\omega \Delta t) = \begin{bmatrix}
        \text{sin}(\frac{1}{2} ||\omega|| \Delta t) \frac{\omega}{||\omega||} \\
        \text{cos}(\frac{1}{2}||\omega|| \Delta t)
    \end{bmatrix} \label{eq:exp_q} ,
\end{align}
$\delta_{\mathbf{x}} \sim \mathcal{N}(0,Q_{\mathbf{x}})$ denotes the process noise. 
\subsubsection{Measurement Model}
Based on the kinematics model \eqref{eq:pfoot}, the Leg Odometry (LO) utilizes the encoder to provide relative position measurements $y_p$ between the base and foot:
\begin{equation}
    y_p = p - p_{\text{foot}} + \delta_{y_p}, \quad \tilde{y}_{p} = - R_{\text{WB}} fk(\tilde{\alpha}), \label{eq:pfootmeas}
\end{equation}
where $\delta_{y_p} = \delta_{\text{pf}}$ denotes the error of the measurement model. Once static contact is established between the foot and the ground, floating base velocity $y_v$ is measured based on \eqref{eq:vfoot}: 
\begin{align}
    y_{v} &= v + \delta_{y_{v}}, \ ( \text{if foot is in contact}) , \label{eq:vfootmeas}\\
    \tilde{y}_{v} &=  - R_{\text{WB}} J(\tilde{\alpha}) \tilde{\dot{\alpha}} - R_{\text{WB}} (\tilde{\omega}- b_{\omega})^{\times} fk(\tilde{\alpha}),
\end{align}
where the measurement noise $\delta_{y_{v}}= \delta_{\text{vf}} + \delta_{\text{slip}}$, and $\delta_{\text{slip}} \sim \mathcal{N}(0,Q_{\text{slip}})$ denotes foot slipping noise. 

\subsection{Visual Odometry}
Visual Odometry (VO) measures the robot pose in the camera frame $C$ by tracking features in the images from onboard cameras. Without loop closure \cite{Campos_2021}, the VO output is interpreted as the incremental homogeneous transformation between consecutive camera frames $C_i$ and $C_j$: $\mathbf{T}_{\text{C}ij} = \mathbf{T}_{\text{WC}_i}^{-1} \mathbf{T}_{\text{WC}_j} + \delta_{\text{vo}}$, 
where $\mathbf{T}_{\text{WC}_i}$ and $\mathbf{T}_{\text{WC}_j}$ are the homogeneous transformations from the world frame $W$ to the camera frame $C$ at time $i$ and $j$. The VO measurement is corrupted by noise $\delta_{\text{vo}} \sim \mathcal{N}(0,Q_{\text{vo}})$. With known fixed transformation $\mathbf{T}_{\text{BC}}$ between IMU/body frame $B$ and camera frame $C$, the transformation of the body frame $\mathbf{T}_{\text{B}_{ij}}$ is represented as:
\begin{equation}
    \tilde{\mathbf{T}}_{\text{B}_{ij}} = \mathbf{T}_{\text{BC}} \tilde{\mathbf{T}}_{\text{C}_{ij}} \mathbf{T}_{\text{BC}}^{-1} . \label{eq:VO_body}
\end{equation}
Typically, the output frequency of the VO matches the camera frame rate, ranging from 10 to 60 Hz. In the state-of-the-art VO implementation, relative transformations are integrated and refined through local or full Bundle Adjustment (BA), yielding the absolute transformation $\tilde{\mathbf{T}}_{\text{WC}_j}$ at time $j$. The absolute transformation $ \tilde{\mathbf{T}}_{\text{WB}_{j}}$ of frame $B$ at time $j$ is:
\begin{equation}
    \tilde{\mathbf{T}}_{\text{WB}_{j}} = \tilde{\mathbf{T}}_{\text{WC}_{j}} \mathbf{T}_{\text{BC}}^{-1} .  \label{eq:VO_abs}
\end{equation}
In our implementation, the initial VO frame is aligned with the IMU frame when the VO is activated, assuming the robot starts in a static pose. In the proposed framework, \eqref{eq:VO_body} will be used to construct state constraints on incremental body displacement in the MHE in Section IV.B, and \eqref{eq:VO_abs} corrects the orientation estimation in the EKF in Section IV.A.

\section{Decentralized estimation Framework}
Considering the nonlinearity of the legged state dynamics and the extensive number of states and constraints involved, the corresponding nonlinear MHE poses significant computational challenges. We thus propose a novel approach to decentralize the nonlinear estimation of the floating base into a nonlinear orientation estimation using EKF and a linear velocity estimation using MHE, both of which can then be solved online. Similar decoupling has been successfully applied for estimation using KFs on the HRP2 humanoid \cite{8246977}, the Cassie biped \cite{Xiong20213DUB}, and MIT Cheetah 3 quadruped \cite{8593885}.
% where similar decoupled schemes were used.

\subsection{Nonlinear Orientation Estimation}
The first part of the decentralized estimation employs an orientation filter. For nine-axis IMU, off-the-shelf AHRS (Attitude and Heading Reference Systems) are able to provide a reasonable orientation measurement $\hat{R}_{\text{WB}}$ relative to the direction of gravity and the earth magnetic field \cite{madgwick2010efficient}. 

In dynamic legged locomotion, the fast changes of accelerations often lead to estimation drifts using off-the-shelf AHRS; also, the magnetometer readings are often disrupted by electromagnetic effects. To address this, the absolute orientation output from VO is used as an additional measurement alongside the accelerometer to improve estimation. An Iterated EKF \cite{Moduler} is used to fuse the IMU gyroscope with the accelerometer readings \cite{s111009182} and VO measurements. 
% While the orientation estimation works for any parameterization of $R_{\text{WB}}$, we use the quaternion representation here. 
\subsubsection{Process Model}
The orientation dynamics is:
\begin{equation}
    \textstyle
    q^+ = (I + \frac{1}{2}\Omega(\omega) \Delta t ) q + W(q)\delta_{\omega},  
\end{equation}
where $I$ is the $4 \times 4$ identity matrix, and $I + \frac{1}{2}\Omega(\omega) \Delta t$ and $W(q)$ are the first-order approximated Jacobian of the exponential map \eqref{eq:exp_q} over the quaternion $q$ and the angular rate $\omega$; for the explicit definitions, see \cite{s111009182}.

\subsubsection{Measurement Model}
The drifts in the roll and pitch axes are corrected by the measurement of the projected gravitational acceleration $\text{g}$ in the body frame $B$ \cite{s111009182}: 
\begin{equation}
    y_{a} = R_{\text{WB}}^T \text{g} + \kappa \delta_{a} , \quad \tilde{y}_{a} = \tilde{a} ,
\end{equation}
where $\kappa = {||\tilde{a}||}/{||\text{g}||}$.
% The roll and pitch drifts due to dynamic motion and the yaw are corrected by the absolute orientation output of VO:
The absolute orientation output of VO provides an additional measurement:
\begin{equation}
    y_{qc}  = q + \delta_{y_{qc}}, \quad \tilde{y}_{qc}  =  \text{rots}(\tilde{\mathbf{T}}_{\text{WB}}) , \label{eq:voorien} 
\end{equation}
where $\tilde{y}_{qc}$ is the orientation part of $\tilde{\mathbf{T}}_{\text{WB}}$, and $\delta_{y_{qc}} \sim \mathcal{N}(0,Q_{y_{qc}}) $ is the VO measurement noise of the orientation. Although the noise may increase in the absolute output of VO, the presence of loop closure (full BA) allows for the selection of a constant covariance during our implementation. 

The latency of the visual information is addressed based on synchronization and trajectory update \cite{Moduler}. With a buffer of stored IMU measurements, when a new visual measurement arrives at time $T_{\text{vo}}$, the image frame is aligned with the corresponding IMU frame at the same time stamp. Correction is applied at this alignment point using \eqref{eq:voorien} with EKF update. Then the subsequent measurements
are re-applied to this updated state from time $T_{\text{vo}}$ to the current time $T$, resulting in an updated state trajectory estimation from time $T_{\text{vo}}$ to $T$. 

\subsection{Linear Position and Velocity Estimation}
The second part of the decentralized estimation utilizes the orientation estimate $\hat{R}_{\text{WB}}$ from Section IV.A, simplifying the process model of the original nonlinear estimation \eqref{eq:state}, \eqref{eq:nonlinear_dyn}. The remaining state is decentralized from the original definition \eqref{eq:state}, denoted again by $\mathbf{x}$ for simplicity: 
\begin{equation}
    \mathbf{x} := \begin{bmatrix}
        p^{\intercal} & v^{\intercal} & p_{\text{foot}}^{\intercal} &   b_{a}^{\intercal}
    \end{bmatrix}^{\intercal} .
\end{equation}
As in \eqref{eq:MHE}, the process model, measurement model, and additional constraints are all encoded as state constraints.
\subsubsection{Process Constraint}
At each discrete time, with the equivalent control input $u =  \hat{R}_{\text{WB}} a + \text{g}$ in \eqref{eq:nonlineardiff}, the state $\mathbf{x}$ evolves on the time-varying linear dynamics:
\begin{equation}
    \mathbf{x}^{+} = \begin{bmatrix}
        p + v \Delta t + \frac{1}{2} \big(\hat{R}_{\text{WB}} (\tilde{a} - b_{a}) + \text{g} \big) \Delta t^2\\
        v + \big(\hat{R}_{\text{WB}} (\tilde{a} - b_{a}) + \text{g}\big) \Delta t\\
        p_{\text{foot}}\\
        b_{a}\\
    \end{bmatrix} + \delta_{\mathbf{x}} , \label{eq:linear_process}
\end{equation}
where $\delta_{\mathbf{x}} \sim \mathcal{N}(0,Q_{\mathbf{x}})$ represents the noise of the dynamics:
\begin{equation}
    \delta_{\mathbf{x}} = \begin{bmatrix}  \hat{R}_{\text{WB}} \delta_{p} + \frac{1}{2} \hat{R}_{\text{WB}} \Delta t^2 \delta_{a} \\
    \hat{R}_{\text{WB}} \Delta t \delta_{a} \\
    \hat{R}_{\text{WB}} \delta_{\text{foot}} \\
    \delta_{b_a}
    \end{bmatrix} ,
\end{equation}
where $\delta_p \sim \mathcal{N}(0,Q_{p})$ is introduced to account for integration inaccuracies. $\delta_{\text{foot}} \sim \mathcal{N}(0,Q_{\text{foot}})$ denotes the foot process noise with large covariance, enabling the relocation of the foot position. At each time index $k$, the process model \eqref{eq:linear_process} composes consecutive state constraints in the MHE:
\begin{equation}
     \mathbf{x}_{k+1} = A_k \mathbf{x}_k + b_k + \delta_{\mathbf{x}_k},\ k\in \{T-N,...,T-1 \}. \tag{Dyn.} \label{eq:Dynamics}
\end{equation}    
\subsubsection{Measurement Constraint}
% The LO model of foot position in \eqref{eq:pfootmeas} and foot velocity in \eqref{eq:vfootmeas} are directly formulated as linear constraints:
The LO model in \eqref{eq:pfootmeas} and \eqref{eq:vfootmeas} are directly formulated as linear constraints:
\begin{align}
    y_{k} &= H_k \mathbf{x}_k + \delta_{y_k} ,\ k \in \{T-N, ... , T \}. \tag{LO}\label{eq:LO}
\end{align}
For simplicity, $y$, $\delta_y$, and $Q_y$ will denote the measurement, error, and covariance of the LO, respectively, and the choice between \eqref{eq:pfootmeas} and \eqref{eq:vfootmeas} depends on the implementation. 

\subsubsection{Physical Constraint}
% \noindent{\textbf{\emph{Remark}}}: In this paper, we model
The foot-ground contact is modeled as a deterministic constraint, i.e., the foot does not slip on the ground. This deterministic state constraint model will further demonstrate the capability of the MHE to handle constraints. When the foot is in stable contact: 
\begin{equation}
    (p_{\text{foot}}^+ - p_{\text{foot}}) f_{\text{foot}}^{\text{grf}} = 0, \label{eq:contact_LCP}
\end{equation}
where $f_{\text{foot}}^{\text{grf}} $ is the ground normal force reacted on the foot; it is equivalent to a boolean variable when using a contact sensor. 
Alternatively, this stationary constraint of foot position can be replaced by the velocity constraint $v_{\text{foot}} f_{\text{foot}}^{\text{grf}} = 0 $. These linear state constraints are denoted as: 
\begin{align}
    f_k = F_{k} \mathbf{x}_k,\ k \in \{T-N,...,T\}. \tag{Contact} \label{eq:contact} 
\end{align}
Note that this is a simplification of the Linear Complementary Problem (LCP) formulation \cite{DBLP:conf/wafr/VarinK18} of rigid body contact. It works well when the control enforces the static foot contact, and eliminates the need for extensive tuning of the sliding covariance. When walking on slippery terrain, the stochastic contact model in EKF \cite{bloesch2013state} can still be enforced in the MHE. 

\begin{figure}[t]
    \centering
    \setlength{\abovecaptionskip}{0pt}
    \includegraphics[width=.99\linewidth]{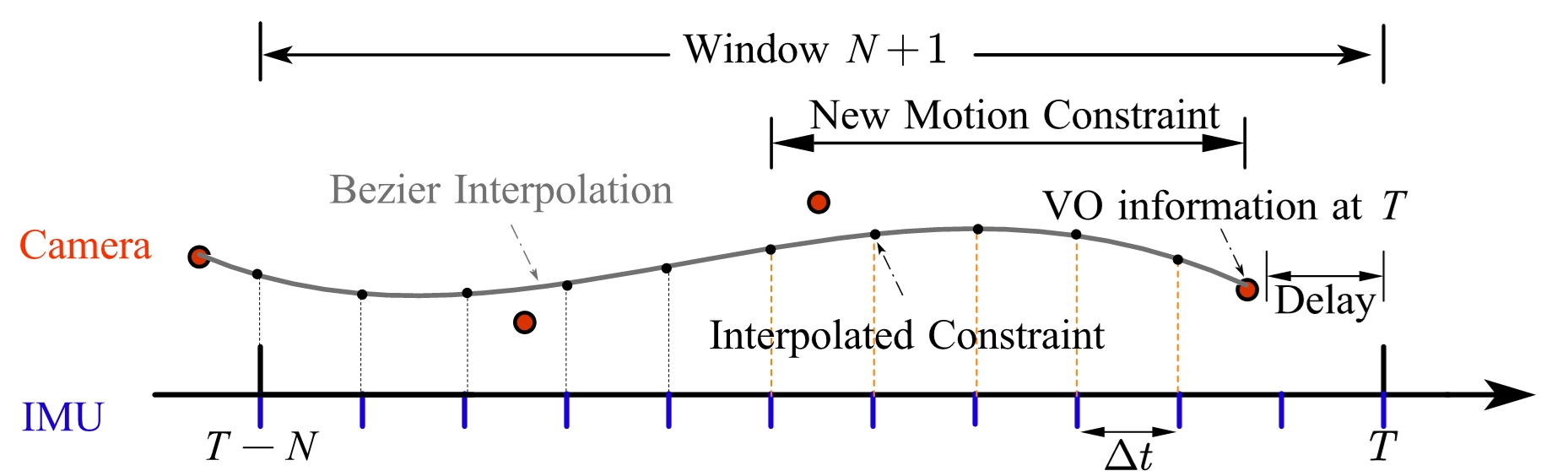}
    % \includesvg[width=.99\linewidth]{figure/interpolation.svg}
        \vspace{-15pt}
    \caption{The VO measurements are synchronized with the nearest IMU frames and interpolated at all IMU frames as the motion constraints.}
    \label{Synchronize}
    \vspace{-18pt}
\end{figure}

\subsubsection{Synchronization and Visual Odometry Constraint}
The proprioceptive and exteroceptive sensors are of different frequencies in general, so synchronization is necessary for sensor fusion. Every camera/VO frame is aligned to the closest IMU frame. After the alignment, only the translation component $p_{c_{ij}}$ of the incremental transformation $\mathbf{T}_{\text{B}_{ij}}$ from \eqref{eq:VO_body} is applied as a measurement into the estimation:
\begin{equation}
        \tilde{p}_{c_{ij}} = \hat{R}_{\text{WB}_i} \text{trans}(\tilde{\mathbf{T}}_{\text{B}_{ij}}) .
\end{equation}
We use a Cubic Bézier polynomial to fit a series of relative translation measurements $\tilde{p}_{c_{ij}}$. As illustrated in Fig. \ref{Synchronize}, a smooth path is generated using consecutive VO translations as control points. The smooth curve then provides measurements between consecutive IMU frames $k$ and $k+1$, $\forall k \in \{k \in \mathbb{N}^+ | \ i \leq k \leq j-1\}$:
\begin{equation}
    y_{c_k} = p_{k+1} - p_{k} + \delta_{c_k}  , \quad \tilde{y}_{c_k} = \tilde{p}_{c_k} , \label{eq:cam_measure}
\end{equation}
where $\tilde{p}_{c_k}$ is the smoothed output of Cubic Bézier interpolation between consecutive IMU frames $k$ and $k+1$, and $\delta_{c_k}$ denotes the VO measurement noise. Using the most recent VO measurement with a delay at time index $T_{\text{vo}}$, the VO constraints are formulated as:
\begin{equation}
    y_{c_k} = C \mathbf{x}_k - C \mathbf{x}_{k+1} + \delta_{c_k},\ k \in \{T-N,...,T_{\text{vo}}-1\}. \tag{VO} \label{eq:VO} 
\end{equation}
\subsubsection{Linear MHE}
With the linear constraints constructed above, the \eqref{eq:MHE} is formulated as a constrained QP:
\begin{align}
 {\hat{\mathcal{X}}_{[T-N,T]}} =  & \ \underset{\mathcal{X}_{[T-N,T]}}{ \text{argmin}}  \ J(\mathcal{X}_{[T-N,T]}) \label{eq:QP},\\ 
\text{s.t.} \quad & \eqref{eq:Dynamics}, \eqref{eq:LO},  \eqref{eq:contact}, \eqref{eq:VO}, \label{eq:QP_constraints}
\end{align}
where the optimization variable is defined as:
\begin{small}
\begin{align*}
        \mathcal{X}_{[T-N,T]} & = \begin{bmatrix}
        \mathbf{x}_{[T-N,T]} \ \delta_{x_{[T-N,T-1]}} \ \delta_{c_{[T-N,T_{\text{vo}}-1]}} \ \delta_{y_{[T-N,T]}} 
    \end{bmatrix}, 
    % ,\\
    %      \delta_{c_{[T-N,T_{\text{vo}}-1]}} & = \begin{bmatrix}
    %   \delta_{c_{T-N}} & \delta_{c_{T-N+1}} & ... & \delta_{c_{T_{\text{vo}}-1}}
    % \end{bmatrix} , \\
    %   \delta_{y_{[T-N,T]}} & = \begin{bmatrix}
    %   \delta_{y_{T-N}} & \delta_{y_{T-N+1}} & ... & \delta_{y_{T}} \\
    % \end{bmatrix} , 
\end{align*}    
\end{small}
and the objective function is: 
\begin{align}
   % \begin{split}
         J(\mathcal{X}_{[T-N,T]}) =  \Gamma(\mathcal{X}_{T-N}) + \textstyle \sum_{k=T-N}^{T-1}{\delta_{\mathbf{x}_k}^\intercal Q^{-1}_{\mathbf{x}_k} \delta_{\mathbf{x}_k}} \nonumber \\
        + \textstyle \sum_{k = T-N}^T{\delta_{y_k}^\intercal Q^{-1}_{y_k} \delta_{y_k}} + \textstyle \sum_{k = T-N}^{T_{\text{vo}}-1}{\delta_{c_k}^\intercal Q^{-1}_{c_k} \delta_{c_k}},
    %\end{split} 
    \label{eq:MHE_J}
\end{align}
where $\mathcal{X}_{T-N} = [\mathbf{x}_{T-N}\ \delta_{\mathbf{x}_{T-N}}\ \delta_{c_{T-N}}\ \delta_{y_{T-N}}]$.
The MHE involves all measurements of different frequencies within the window. The calculation of $\Gamma(\mathcal{X}_{T-N})$ is explained next.

% The delayed VO measurements are appended as constraints, as a result, the trajectory update \cite{Moduler} in Section IV.A is not required to address the latency.    

\section{Arrival Cost Calculation}
With a receding horizon, MHE iteratively marginalizes the oldest
measurements from the window and appends new measurements to it. During this process, the equivalence between MHE and FIF is maintained by a proper choice of arrival cost $\Gamma(\cdot)$. Due the fact that the linear MHEs are Quadratic Programs (QPs), the optimal arrival cost can be calculated based on its optimality condition. %We restrict the discussion to equality-constrained linear MHE. 
\subsection{MHE as Quadratic Program} The equality-constrained linear MHE \eqref{eq:QP} can be written into the generalized form of QP:
% \begin{align}
%  \min_{x}& \  \frac{1}{2}x^\intercal H x + h^\intercal x \label{eq:generalQP},\\ 
% \text{s.t.}& \quad G x = g \nonumber.
% \end{align}
\begin{align}
    \min_{\mathcal{X}_{[T-N,T]}} & \ \textstyle \frac{1}{2} \mathcal{X}_{[T-N,T]}^\intercal \mathbf{H} \mathcal{X}_{[T-N,T]} + \mathbf{h}^{\intercal} \mathcal{X}_{[T-N,T]}  \label{eq:linearMHEobj} \\
 \text{s.t.}& \quad \mathbf{G} \mathcal{X}_{[T-N,T]} = \mathbf{g}. \label{eq:linearMHE}
\end{align}
In this case, $\mathbf{G}$ has full row rank. The optimality condition, i.e., the  Karush–Kuhn–Tucker (KKT) equation, is:
\begin{equation}
     \underset{\mathbf{K}}{\underbrace{\begin{bmatrix}
     \mathbf{H} & \mathbf{G}^\intercal \\
     \mathbf{G} & \mathbf{0}
    \end{bmatrix} }} 
    \begin{bmatrix}
     \mathcal{X}_{[T-N,T]}^* \\ \lambda_{[T-N,T]}^*
    \end{bmatrix} = 
       \underset{\mathbf{k}}{\underbrace{\begin{bmatrix}
     -\mathbf{h} \\ \mathbf{g} 
    \end{bmatrix}}}, \label{eq:KKT}
\end{equation}
where $\lambda_{[T-N,T]} = [\lambda_{T-N},...,\lambda_T]$ is the Lagrange multiplier, and $\mathbf{K}$ is the KKT matrix. Note that this condition is necessary and sufficient, and it is derived from the first-order optimality condition of the Lagrangian of the QP: 
\begin{align*}
    L = J(\mathcal{X}_{[T-N,T]})+ \lambda_{[T-N,T]}^\intercal (\mathbf{G}\mathcal{X}_{[T-N,T]} - \mathbf{g}).
\end{align*}
% \begin{equation}
%     L = \frac{1}{2}x^\intercal H x + h^\intercal x + \lambda^\intercal (Gx - g).
% \end{equation}
Directly solving the KKT equation \eqref{eq:KKT} can be numerically challenging especially when $\mathbf{H}$ and $\mathbf{G}$ are of large dimensions. Off-the-shelf solvers have different numerical procedures to solve the problem iteratively. Nevertheless, the optimality condition is used to determine the optimal arrival cost function for the MHE problem. 

\subsection{Arrival Cost Computation}
We present the computation of arrival cost at time $T = N + 1$; the consecutive computation follows the example with the window shifting forward. When $T \leq N$, the MHE is equivalent to FIF, and the arrival cost is equal to the prior cost \eqref{eq:arrival0}, which can be written as $\Gamma(\mathcal{X}_0) = \frac{1}{2} \mathcal{X}_0^{\intercal} M_0 \mathcal{X}_0 + m_0^\intercal \mathcal{X}_0$. At $T = N+1$, the \textit{pre-marginalized MHE}, that is the ``previous MHE" problem at $T-1$ with additional measurements at current time $T$, can be written as:
\begin{align}
    \min_{\mathcal{X}_{[0,T]}} & \ J(\mathcal{X}_{[0,T]}), \label{eq:linearFIobj} \\
 \text{s.t.}& \quad \mathbf{G} \mathcal{X}_{[0,T]} = \mathbf{g}, \label{eq:linearFI}
\end{align}
which has a window size $N+2$; it is equivalent to the FIF. $J(\mathcal{X}_{[0,T]})$ is the quadratic cost of the \textit{pre-marginalized MHE} that can be written into the generalized form as in \eqref{eq:linearMHEobj}:
\begin{equation}
\textstyle
    J(\mathcal{X}_{[0,T]}) = \frac{1}{2} \mathcal{X}_{[0,T]}^\intercal \mathbf{H} \mathcal{X}_{[0,T]} + \mathbf{h}^{\intercal} \mathcal{X}_{[0,T]} .
\end{equation}
The oldest optimization variable $\mathcal{X}_0 = [\mathbf{x}_0\ \delta_{\mathbf{x}_0}\ \delta_{c_0}\ \delta_{y_0}]$ should be marginalized from \eqref{eq:linearFIobj} to maintain a fixed window size of the MHE \eqref{eq:QP}, and the related constraints of $\mathcal{X}_0$ are incorporated to the arrival cost $\Gamma(\mathcal{X}_{1})$. 
Based on the optimization structure of \eqref{eq:linearFIobj}, it can be rewritten as:
% From an optimization perspective \cite{5980267}, establishing this equivalency involves decomposing the FIF \eqref{eq:linearFIobj}, \eqref{eq:linearFI} as:
\begin{align}
  & \min_{\mathcal{X}_1,\mathcal{X}_{[2,T]}}  \  \big(J_r(\mathcal{X}_1,\mathcal{X}_{[2,T]}) 
    + \min_{\mathcal{X}_0}J_m(\mathcal{X}_0,\mathcal{X}_1)\big), \label{eq:decouple_QP}\\
\text{s.t.} &  \underset{\mathbf{G}}{\underbrace{\begin{bmatrix}
    G_{0,0} & G_{0,1}  & 0 \\
    0 & G_{1,1}  & G_{1,[2,T]} \\
    0 & 0  & G_{[2,T],[2,T]}
\end{bmatrix}}} \underset{\mathcal{X}_{[0,T]}}{\underbrace{\begin{bmatrix}
    \mathcal{X}_0\\
    \mathcal{X}_1\\
    \mathcal{X}_{[2,T]}\\
\end{bmatrix} }}=\underset{\mathbf{g}}{\underbrace{ \begin{bmatrix}
    g_0 \\ g_1 \\ g_{[2,T]}
\end{bmatrix}}}, \label{eq:decouple_constraints}
\end{align}
where $J_m(\mathcal{X}_0,\mathcal{X}_1)$ is the part of objective $J(\mathcal{X}_{[0,T]})$ that involves $\mathcal{X}_0$, and $
    J_r(\mathcal{X}_1,\mathcal{X}_{[2,T]}) = J(\mathcal{X}_{[0,T]}) - J_m(\mathcal{X}_0,\mathcal{X}_1)
$. Solving the sub-optimization problem:
\begin{align}
 \underset{\mathcal{X}_0}{\text{min}}  \ &J_m  (\mathcal{X}_0,\mathcal{X}_1\big),  \label{eq:sub_opt}\\
\text{s.t.}\quad  & G_{0,0}
    \mathcal{X}_0  + G_{0,1} \mathcal{X}_1 = g_0,
 \label{eq:subopt}
\end{align}
results in the ideal arrival cost $\Gamma(\mathcal{X}_1)^*$ in the \textit{current MHE} \eqref{eq:QP}. This also eliminates the first row of constraints in \eqref{eq:decouple_constraints}, therefore, converting \eqref{eq:decouple_QP} to the \textit{current MHE} \eqref{eq:QP}. The closed-form solution of this arrival cost $\Gamma(\mathcal{X}_1)$ is calculated using the KKT equations of the \textit{pre-marginalized MHE} \eqref{eq:linearFIobj}:

\begin{equation}
   \underset{\mathbf{K}}{\underbrace{ \begin{bmatrix}\begin{array}{c|ccc}
        K_{0,0} & K_{0,1} & 0 & 0 \\ \hline
        K_{1,0} & K_{1,1} & ... & ... \\
        0 & ... & ... & ...\\
        0 & ... & ... & ...\\
        \end{array}
    \end{bmatrix} }} 
    \underset{\mathcal{L}_{[0,T]}}{\underbrace{\begin{bmatrix}
        \mathcal{L}_0 \\
        \mathcal{X}_1 \\
        \lambda_1\\
        \mathcal{L}_{[2,T]}
    \end{bmatrix}}} =  \underset{\mathbf{k}}{\underbrace{\begin{bmatrix}
        k_0 \\
        k_1 \\
        ... \\
        ...
    \end{bmatrix}}} , \label{eq:KKT_QP_reorder}
\end{equation}
where $\mathcal{L} = \begin{bmatrix}
    \mathcal{X} & \lambda
\end{bmatrix}$; \eqref{eq:KKT_QP_reorder} is reordered from \eqref{eq:KKT} w.r.t. the order of $\mathcal{L}_{[0,T]}$.
Solving the Schur complement of the top-left corner of the KKT matrix $\mathbf{K}$ w.r.t. $\mathcal{L}_0$ yields the KKT equation of the \textit{current MHE} \eqref{eq:QP} that is marginalized from the \textit{pre-marginalized MHE} \eqref{eq:linearFIobj} \cite{FRISON201580}. Converting the resultant KKT equation back to the associated QP, that is, the \textit{current MHE} in the form of \eqref{eq:QP}, we get the closed-form expression of its arrival cost $\Gamma(\mathcal{X}_1)$:
\begin{align}
        \Gamma(\mathcal{X}_1) &= \textstyle \frac{1}{2} \mathcal{X}_1^{\intercal} M_1 \mathcal{X}_1 + m_1^\intercal \mathcal{X}_1 , \label{eq:arrival_closed}\\
    M_1 &= K_{1,1} - K_{1,0} K_{0,0}^{-1} K_{0,1} , \label{eq:arrival_hessian}\\
    m_1 &= - k_1 + K_{1,0} K_{0,0}^{-1} k_0 , \label{eq:arrival_gradient}
\end{align}
which does not require solving the KKT equation \eqref{eq:KKT_QP_reorder} or the \textit{pre-marginalized MHE} \eqref{eq:linearFIobj}.
 The consecutive marginalization process at a new time step follows the example at time $T = N+1$: the closed-form expression of $\Gamma(\mathcal{X}_{T-N})$ in \eqref{eq:QP} at time $T$ is calculated using \eqref{eq:arrival_closed}, with \eqref{eq:arrival_hessian} and \eqref{eq:arrival_gradient} obtained from the KKT condition of the \textit{pre-marginalized MHE}. For the equality-constrained linear MHE, the above formulation entails no approximation, resulting in an MHE maintaining the optimality of FIF. The marginalization is also computationally efficient as it only requires the inversion of $K_{0,0}$. With the arrival cost computed in closed form, any QP solver can be used to solve the MHE problem online.

\begin{figure}[b]
    \centering
    \setlength{\abovecaptionskip}{0pt}
    \includegraphics[width=1.0\linewidth]{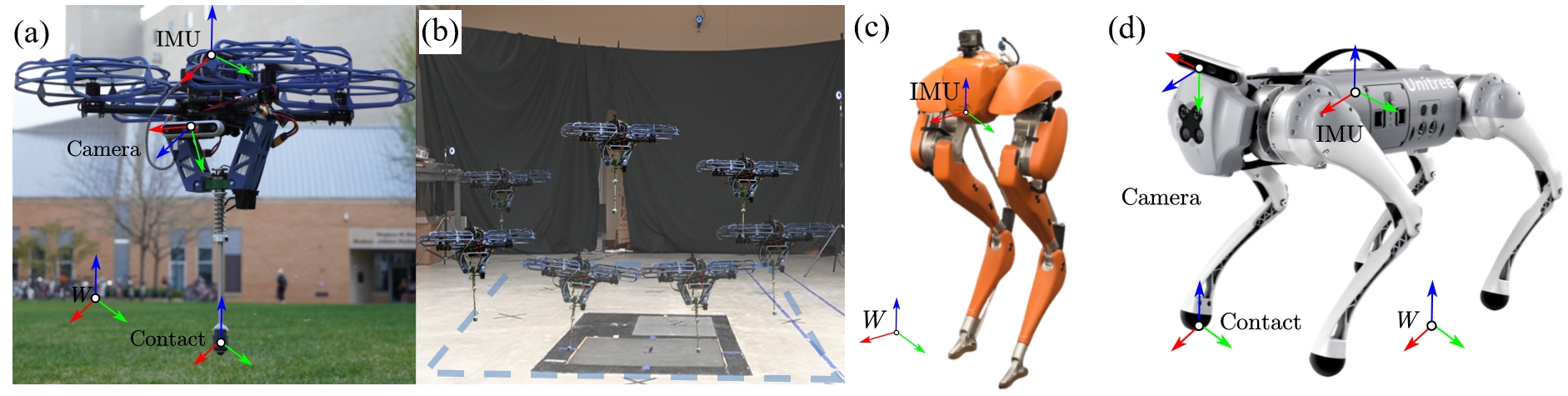}
    \vspace{-5pt}
    \caption{(a) PogoX hardware and reference frames. (b) Indoor experiment of square hopping. (c) Bipedal robot Cassie hardware and reference frames. (d) Quadrupedal robot Unitree Go1 hardware and reference frames.}
    \label{Robot}
    \vspace{-10pt}
\end{figure}

\begin{figure}[t]
    \centering
    \setlength{\abovecaptionskip}{0pt}
    \includegraphics[width=.95\linewidth]{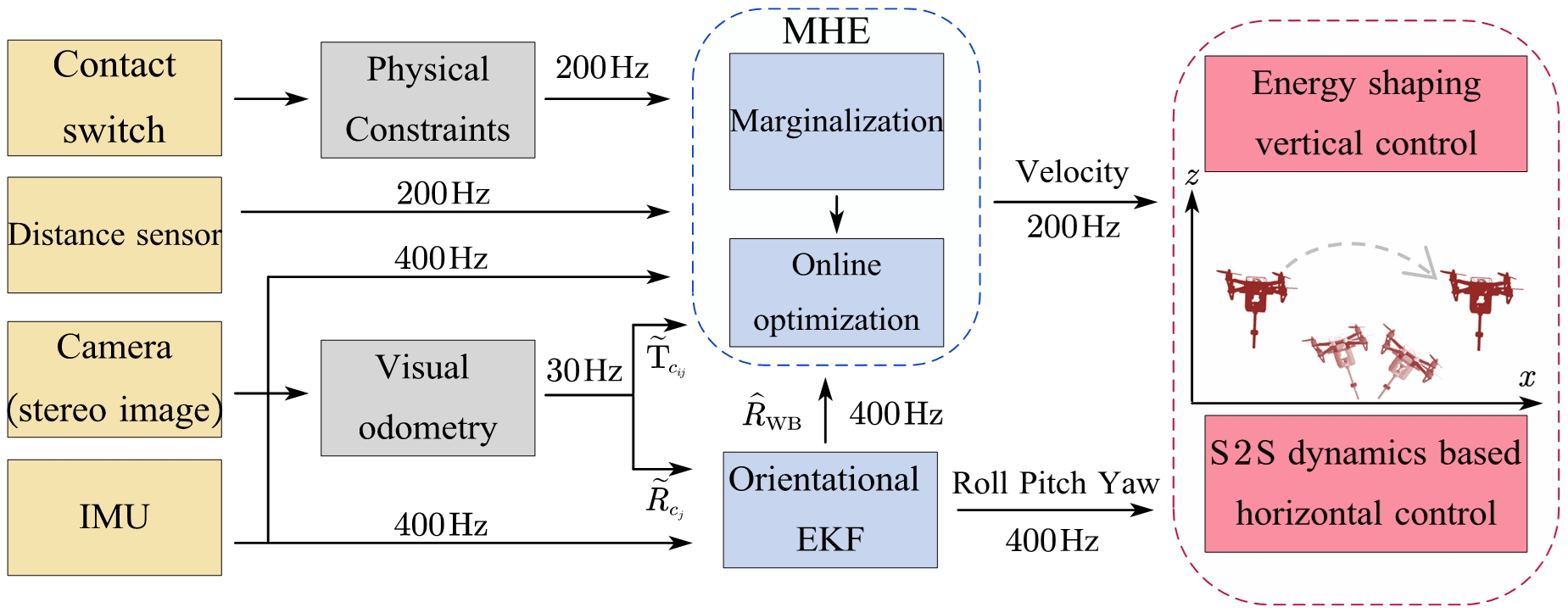}
    % \includesvg[width=1.\linewidth]{figure/pogox_block.svg}
    \vspace{-5pt}
    \caption{Block Diagram of the Decentralized Estimation on PogoX.}
    \vspace{-18pt}
    \label{block_pic}
\end{figure}

\section{Evaluation}
Now we evaluate the proposed decentralized estimation algorithm on our custom-designed hopping robot PogoX, the commercial quadrupedal robot Unitree Go1 and the bipedal robot Cassie. Both the EKF and MHE are implemented using C++ in ROS2 environment. The MHE is solved using OSQP \cite{osqp}, and the VO is implemented via the open-source ORB-SLAM3 \cite{Campos_2021}. The MHEs are solved at 200 Hz, with a window size of 20 and 3-4 camera frames contained, without extensive effort on optimizing the implementation. The software implementation is open-sourced at \cite{Github}. For indoor testing, we use a motion-capture system consisting of 12 OptiTrack Prime-13s to obtain the ground truth at 200Hz. In legged robot estimation, yaw is typically unobservable, so velocity estimates in the following sections are given in the body frame rather than the world frame. 

% \subsection{Hopping Robot PogoX}
\noindent{\textbf{Hopping Robot PogoX:}}
We demonstrate our state estimation algorithm on the robot PogoX developed in \cite{wang2023terrestrial}, which is shown in  Fig. \ref{Robot} (a). We upgraded the mechanical design and the sensors of the original PogoX. This new version of PogoX weighs 3.65 kg, with a height of 0.53 m, and it has a quadcopter frame with a diagonal axle distance of 500 mm. The robot is equipped with a Garmin LIDAR-Lite v3HP LED distance measurement sensor, with an accuracy of ±5 cm and an update rate of 200 Hz. It is used for measuring the distance between the robot and the ground, equivalent to the leg kinematics on other legged robots. We use the Intel Realsense Depth Camera D455 for the onboard vision and an NGIMU with a gyroscope range of 2000 deg/s and an accelerometer sensitivity of 16 gs for inertial measurement. All state estimations are performed in real-time on an onboard Intel NUC 13 Pro (i7 13th Gen 1340P, P-core 5.00 GHz, E-core 3.70 GHz). Fig. \ref{block_pic} shows the block diagram of the estimation. 

\begin{figure}[t]
    \centering
    \setlength{\abovecaptionskip}{0pt}
    \includegraphics[width=.99\linewidth]{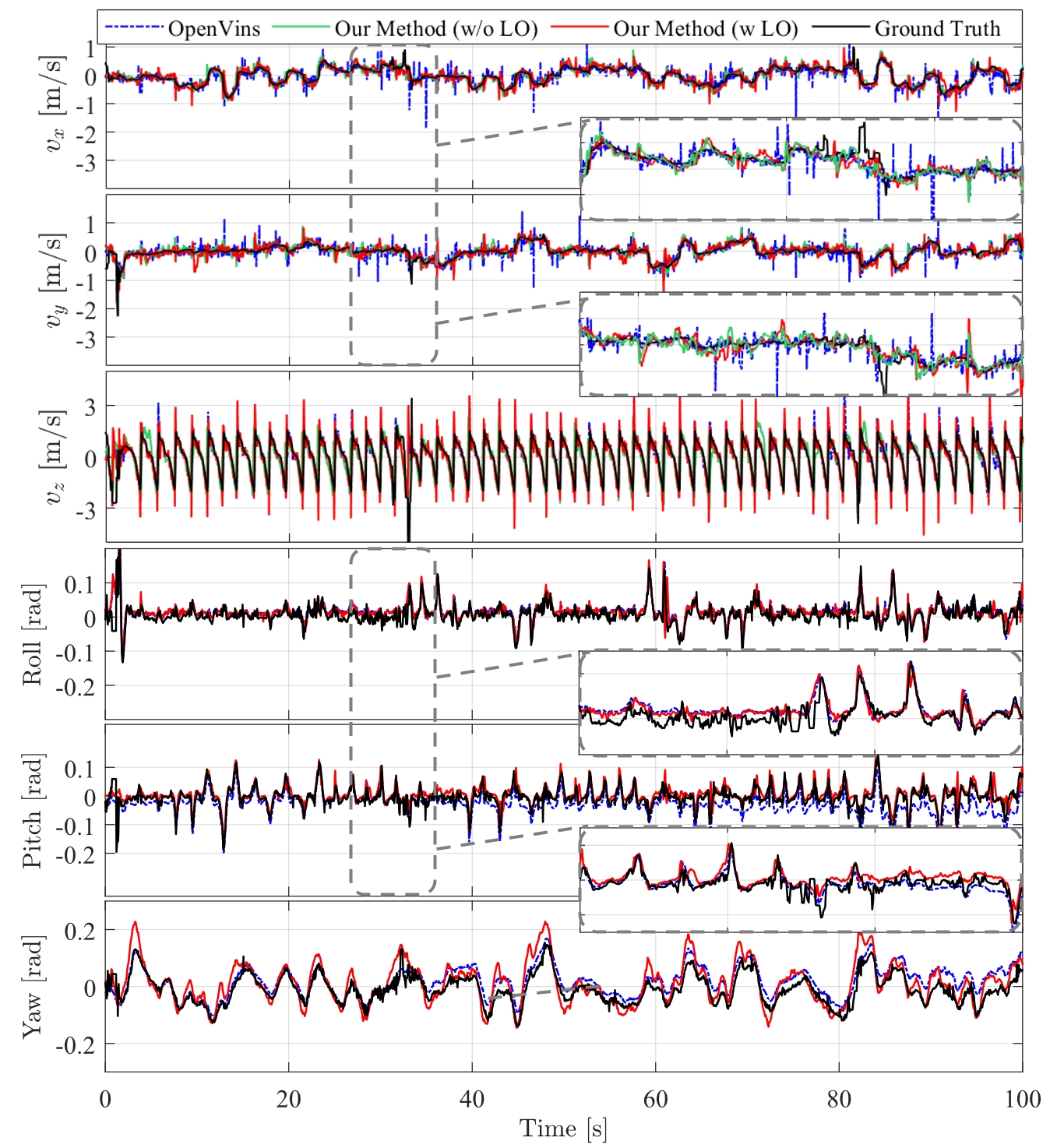}
    \vspace{-15pt}
    \caption{Online velocity and orientation estimation of PogoX experiment. The RMSEs of ours and OpenVINS for horizontal velocity are 0.1351 m/s and 0.1380 m/s. The RMSEs for orientation are 0.0424 rad and 0.0771 rad. Here the LO implies the distance LIDAR. }
        \vspace{-10pt}
    \label{Indoor}
\end{figure}

Snapshots of indoor experiments are presented in Fig. \ref{Robot} (b). The robot is controlled to follow a square path in a clockwise direction. Fig. \ref{Indoor} presents the estimated robot velocity and orientation. The results show that our state estimation closely matches the ground truth. Compared with the widely used VIO OpenVINS \cite{9196524}, where the same noise covariances are used, the RMSEs of our method are reasonably better. Note that the distance LIDAR does not contribute sufficiently to the velocity estimation due to its low precision measurements, yet it is necessary to include to estimate the vertical height that is used for hopping control. Including the distance LIDAR reduced the RMSE of height estimation from 1.1095m to 0.0329m. The brief ground truth outages in Fig. \ref{Indoor} are caused by marker occlusion. For outdoor experiments, our robot can stably perform omnidirectional dynamic hopping on hard ground and soft grass, relying only on the onboard sensors and computation. This shows that our algorithm provides accurate and fast online state estimation.

\begin{figure}[t]
    \centering
    \setlength{\abovecaptionskip}{0pt}
    \includegraphics[width=.99\linewidth]{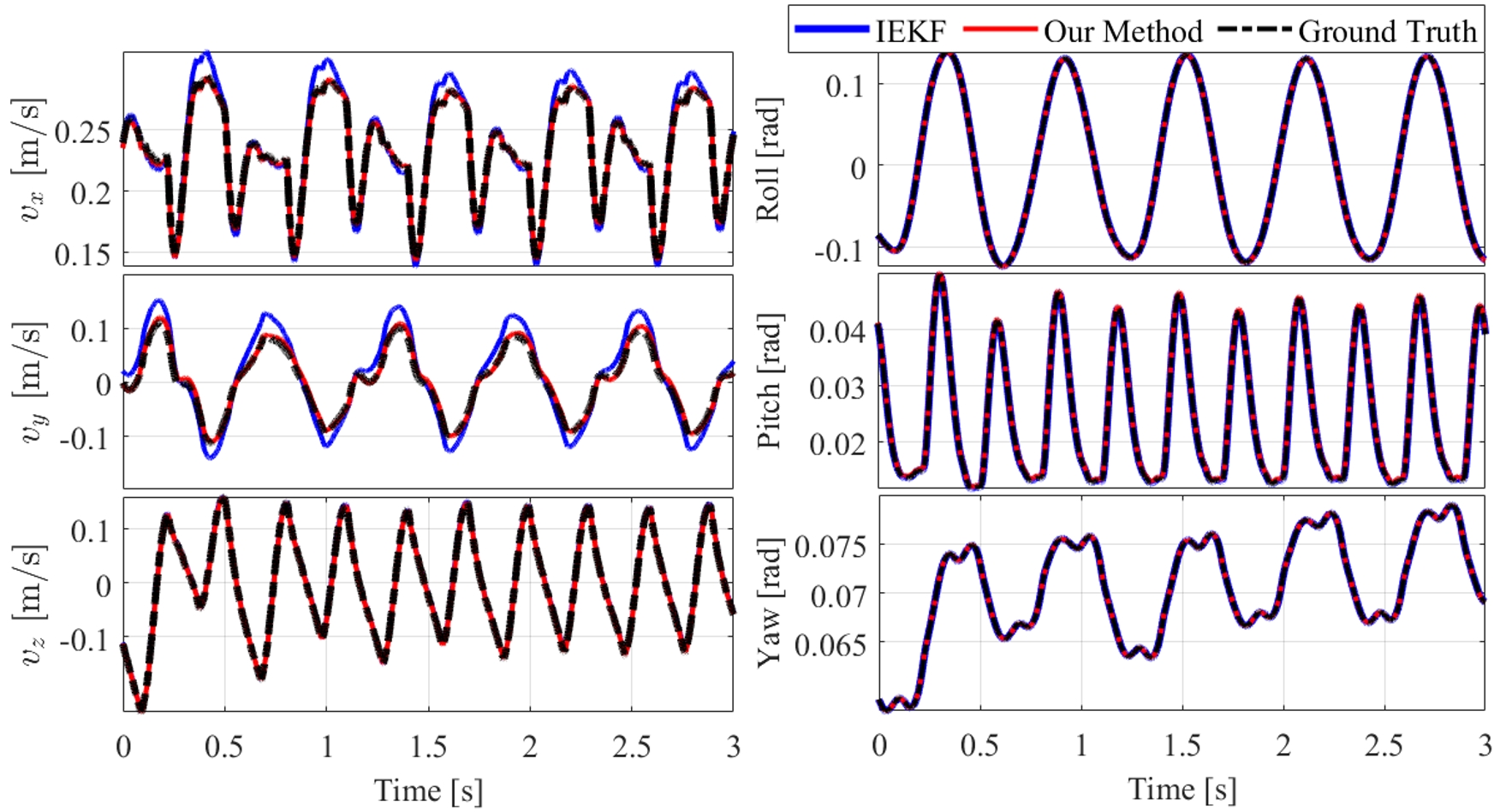}
    \vspace{-15pt}
    \caption{Estimation results on Cassie. The RMSEs of the linear velocities of our method and IEKF are 0.0113 m/s and 0.0283 m/s, respectively.}
    \label{Cassie}
        \vspace{-13pt}
\end{figure}

\begin{figure}[t]
    \centering
    \includegraphics[width=1.0\linewidth]{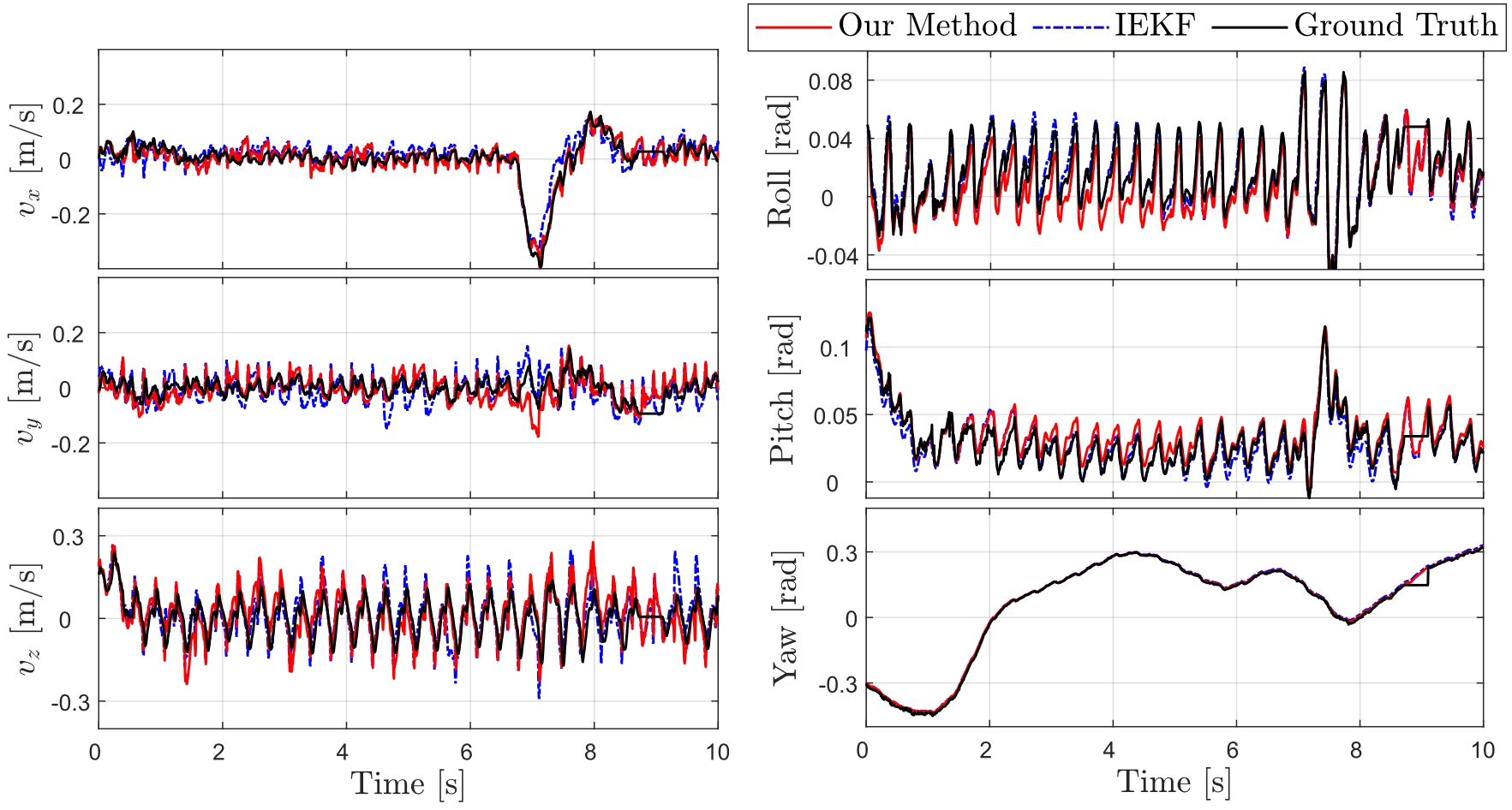}
    % \includesvg[width=1.0\linewidth]{figure/go1_hardware.svg}
    \vspace{-20pt}
    \caption{Estimation results on Unitree Go1 hardware.}
    \label{Go1}
    \vspace{-18pt} % don't use it too aggressively
\end{figure}

\noindent{\textbf{Bipedal Cassie and Quadrupedal Unitree Go1:}}
We evaluate our algorithms on Cassie and Go1. Our method shows better performance than the state-of-the-art estimation algorithms. We first validate our estimation on the simulation dataset \cite{hartley2019contactaided} of Cassie. Fig. \ref{Cassie} shows the comparison result between our method, IEKF\cite{hartley2019contactaided}, and the ground truth. Using the same sensors and noise parameters, our method performs better than IEKF. We also test our estimation algorithms on Go1. During the experiment, we attach an additional Realsense camera D455 to the torso. Fig. \ref{Go1} and Table. \ref{table:RMSE} show that our method closely matches the ground truth, and achieves better performance compared with EKF \cite{bloesch2013state} and IEKF \cite{hartley2019contactaided} that fuse IMU and Leg Odometry.
 
\begin{figure}[t]
    \centering
    % \includesvg[width=0.85\linewidth]{figure/RMSE_window.svg}
     \includegraphics[width=0.80\linewidth]{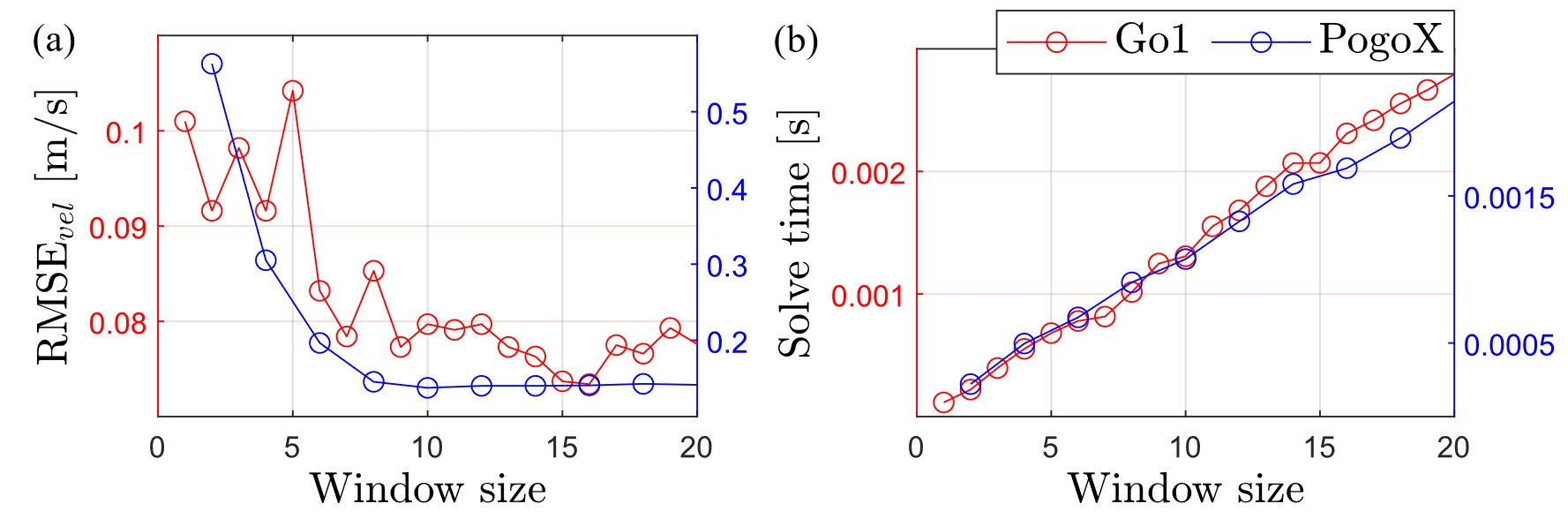}
    \vspace{-8pt}
 \caption{Window size effects on (a) estimation RMSE of linear velocities and (b) computation cost.}
    \label{window}
    \vspace{-15pt} % don't use it too aggressively
\end{figure}

\noindent{\textbf{MHE Window Size Analysis:}}
We investigate the MHE error and computation time in Fig. \ref{window} by increasing the window size from 1 to 20, on both Go1 and PogoX. The RMSE improves with larger windows due to the inclusion of more VO frames; yet errors can rise as VO measurements are discontinuous, especially around the window size of 5. As window size increases beyond 10, the RMSE becomes insensitive to the increase in windows, as sufficient VO frames have been included already. This result highlights the value of MHE in using windowed measurements for precise estimation.
\begin{table}[t]
    \centering
    \caption{RMSEs of the estimations on Go1 hardware.}
     \vspace{-10pt} % don't use it too aggressively
    \begin{tabular}{|>{\centering\arraybackslash}p{1.9cm}|>{\centering\arraybackslash}p{1.4cm}|>{\centering\arraybackslash}p{1.4cm}|>{\centering\arraybackslash}c|>{\centering\arraybackslash}c|}
    \hline
    Estimation & \multirow{2}{*}{Our Method} & Our Method & \multirow{2}{*}{EFK} & \multirow{2}{*}{IEKF} \\
    Method & & (w/o VO) & & \\
    \hline
    $\text{RMSE}_v$ [m/s]  & 0.0654  &  0.0847   &    0.0901 &  0.0817\\
    \hline
    $\text{RMSE}_\text{Euler}$ [rad]& 0.0160  &  0.0230 & 0.0218  &    0.0157 \\
    \hline
    \end{tabular}
    \label{table:RMSE}
    \vspace{-5pt} % don't use it too aggressively
\end{table}

\section{Conclusions and Future Work}
This paper proposes a fast and decentralized state estimation framework based on the Extended Kalman Filter (EKF) and Moving Horizon Estimation (MHE) for the control of legged robots. The original estimation problem is decentralized to an EKF-based orientation estimation and a windowed velocity estimation via MHE. The arrival cost is calculated by the proposed marginalization method, preserving the optimality of the original Full Information Filter. The framework provides a modular approach to the fusion of both proprioceptive and exteroceptive sensor inputs along with physical constraints. Experiments conducted across various legged robots demonstrate that the decentralized framework outperforms state-of-the-art estimation algorithms.

In the future, we aim to exploit the characteristics of QP that have a closed-form solution by utilizing numerical techniques to accelerate computation. We will also conduct theoretical and experimental comparisons on arrival cost calculation between our method and other existing approaches. Additionally, we are interested in exploring different orientation estimators and leveraging windowed optimization to estimate robot states beyond the floating base state. By incorporating physical constraints, we seek to extend the framework to include contact and joint torque estimation in general robots during locomotion and manipulation tasks.

%%%%%%%%%%%%%%%%%%%%%%%%%%%%%%%%%%%%%%%%%%%%%%%%%%%%%%%%%%%%%%%%%%%%%%%%%%%%%%%%

%%%%%%%%%%%%%%%%%%%%%%%%%%%%%%%%%%%%%%%%%%%%%%%%%%%%%%%%%%%%%%%%%%%%%%%%%%%%%%%%

%%%%%%%%%%%%%%%%%%%%%%%%%%%%%%%%%%%%%%%%%%%%%%%%%%%%%%%%%%%%%%%%%%%%%%%%%%%%%%%%

%%%%%%%%%%%%%%%%%%%%%%%%%%%%%%%%%%%%%%%%%%%%%%%%%%%%%%%%%%%%%%%%%%%%%%%%%%%%%%%%
% \newpage
% \addtolength{\textheight}{-5.0cm}
\bibliographystyle{IEEEtran}
\bibliography{reference}

\begin{thebibliography}{10}
\providecommand{\url}[1]{#1}
\csname url@samestyle\endcsname
\providecommand{\newblock}{\relax}
\providecommand{\bibinfo}[2]{#2}
\providecommand{\BIBentrySTDinterwordspacing}{\spaceskip=0pt\relax}
\providecommand{\BIBentryALTinterwordstretchfactor}{4}
\providecommand{\BIBentryALTinterwordspacing}{\spaceskip=\fontdimen2\font plus
\BIBentryALTinterwordstretchfactor\fontdimen3\font minus \fontdimen4\font\relax}
\providecommand{\BIBforeignlanguage}[2]{{%
\expandafter\ifx\csname l@#1\endcsname\relax
\typeout{** WARNING: IEEEtran.bst: No hyphenation pattern has been}%
\typeout{** loaded for the language `#1'. Using the pattern for}%
\typeout{** the default language instead.}%
\else
\language=\csname l@#1\endcsname
\fi
#2}}
\providecommand{\BIBdecl}{\relax}
\BIBdecl

\bibitem{7758092}
M.~Hutter, C.~Gehring, D.~Jud, A.~Lauber, C.~D. Bellicoso, V.~Tsounis, J.~Hwangbo, K.~Bodie, P.~Fankhauser, M.~Bloesch, R.~Diethelm, S.~Bachmann, A.~Melzer, and M.~Hoepflinger, ``Anymal - a highly mobile and dynamic quadrupedal robot,'' in \emph{2016 IEEE/RSJ International Conference on Intelligent Robots and Systems (IROS)}, pp. 38--44.

\bibitem{bloesch2013state}
M.~Bl{\"o}sch, M.~Hutter, M.~A. H{\"o}pflinger, S.~Leutenegger, C.~Gehring, C.~D. Remy, and R.~Y. Siegwart, ``State estimation for legged robots - consistent fusion of leg kinematics and imu,'' in \emph{Robotics: Science and Systems}, 2012.

\bibitem{Wisth_2020}
D.~Wisth, M.~Camurri, and M.~F. Fallon, ``Preintegrated velocity bias estimation to overcome contact nonlinearities in legged robot odometry,'' \emph{2020 IEEE International Conference on Robotics and Automation (ICRA)}, pp. 392--398.

\bibitem{hartley2019contactaided}
R.~Hartley, M.~Ghaffari, R.~M. Eustice, and J.~W. Grizzle, ``Contact-aided invariant extended kalman filtering for robot state estimation,'' \emph{The International Journal of Robotics Research}, vol.~39, no.~4, pp. 402--430, 2020.

\bibitem{6942674}
N.~Rotella, M.~Bloesch, L.~Righetti, and S.~Schaal, ``State estimation for a humanoid robot,'' in \emph{2014 IEEE/RSJ International Conference on Intelligent Robots and Systems}, pp. 952--958.

\bibitem{10.3389/frobt.2020.00068}
M.~Camurri, M.~Ramezani, S.~Nobili, and M.~Fallon, ``Pronto: A multi-sensor state estimator for legged robots in real-world scenarios,'' \emph{Frontiers in Robotics and AI}, vol.~7, 2020.

\bibitem{teng2021legged}
S.~Teng, M.~W. Mueller, and K.~Sreenath, ``Legged robot state estimation in slippery environments using invariant extended kalman filter with velocity update,'' \emph{2021 IEEE International Conference on Robotics and Automation (ICRA)}, pp. 3104--3110.

\bibitem{8594448}
J.~Di~Carlo, P.~M. Wensing, B.~Katz, G.~Bledt, and S.~Kim, ``Dynamic locomotion in the mit cheetah 3 through convex model-predictive control,'' in \emph{2018 IEEE/RSJ International Conference on Intelligent Robots and Systems (IROS)}, pp. 1--9.

\bibitem{6942679}
X.~Xinjilefu, S.~Feng, and C.~G. Atkeson, ``Dynamic state estimation using quadratic programming,'' in \emph{2014 IEEE/RSJ International Conference on Intelligent Robots and Systems}, pp. 989--994.

\bibitem{BAE20173793}
H.~Bae, J.~Oh, H.~Jeong, and J.-H. Oh, ``A new state estimation framework for humanoids based on a moving horizon estimator,'' \emph{IFAC-PapersOnLine}, vol.~50, no.~1, pp. 3793--3799, 2017, 20th IFAC World Congress.

\bibitem{5980267}
T.-C. Dong-Si and A.~I. Mourikis, ``Motion tracking with fixed-lag smoothing: Algorithm and consistency analysis,'' in \emph{2011 IEEE International Conference on Robotics and Automation}, pp. 5655--5662.

\bibitem{wang2023terrestrial}
Y.~Wang, J.~Kang, Z.~Chen, and X.~Xiong, ``Terrestrial locomotion of pogox: From hardware design to energy shaping and step-to-step dynamics based control,'' \emph{2023 IEEE International Conference on Robotics and Automation (ICRA)}.

\bibitem{UNTREEWebsite}
Unitree Go1: \href{https://www.unitree.com/cn/go1}{\texttt{https://www.unitree.com/cn/go1}}.

\bibitem{Github}
\href{https://github.com/well-robotics/Decentralized_EKF_MHE}{\texttt{github.com/well-robotics/Decentralized\_EKF\_MHE}}.

\bibitem{9196524}
P.~Geneva, K.~Eckenhoff, W.~Lee, Y.~Yang, and G.~Huang, ``Openvins: A research platform for visual-inertial estimation,'' in \emph{2020 IEEE International Conference on Robotics and Automation (ICRA)}, pp. 4666--4672.

\bibitem{Hartley_2018}
R.~Hartley, M.~G. Jadidi, L.~Gan, J.-K. Huang, J.~W. Grizzle, and R.~M. Eustice, ``Hybrid contact preintegration for visual-inertial-contact state estimation using factor graphs,'' in \emph{2018 IEEE/RSJ International Conference on Intelligent Robots and Systems (IROS)}.

\bibitem{1178905}
C.~Rao, J.~Rawlings, and D.~Mayne, ``Constrained state estimation for nonlinear discrete-time systems: stability and moving horizon approximations,'' \emph{IEEE Transactions on Automatic Control}, vol.~48, no.~2, pp. 246--258, 2003.

\bibitem{FIF}
C.~K. Enders, ``The performance of the full information maximum likelihood estimator in multiple regression models with missing data,'' \emph{Educational and Psychological Measurement}, vol.~61, no.~5, pp. 713--740, 2001.

\bibitem{Campos_2021}
C.~Campos, R.~Elvira, J.~J.~G. Rodriguez, J.~M. M.~Montiel, and J.~D.~Tardos, ``Orb-slam3: An accurate open-source library for visual, visual–inertial, and multimap slam,'' \emph{IEEE Transactions on Robotics}, vol.~37, no.~6, p. 1874–1890, Dec. 2021.

\bibitem{8246977}
T.~Flayols, A.~Del~Prete, P.~Wensing, A.~Mifsud, M.~Benallegue, and O.~Stasse, ``Experimental evaluation of simple estimators for humanoid robots,'' in \emph{2017 IEEE-RAS 17th International Conference on Humanoid Robotics (Humanoids)}, pp. 889--895.

\bibitem{Xiong20213DUB}
X.~Xiong and A.~Ames, ``3-d underactuated bipedal walking via h-lip based gait synthesis and stepping stabilization,'' \emph{IEEE Transactions on Robotics}, vol.~38, pp. 2405--2425, 2021.

\bibitem{8593885}
G.~Bledt, M.~J. Powell, B.~Katz, J.~Di~Carlo, P.~M. Wensing, and S.~Kim, ``Mit cheetah 3: Design and control of a robust, dynamic quadruped robot,'' in \emph{2018 IEEE/RSJ International Conference on Intelligent Robots and Systems (IROS)}, pp. 2245--2252.

\bibitem{madgwick2010efficient}
S.~Madgwick \emph{et~al.}, ``An efficient orientation filter for inertial and inertial/magnetic sensor arrays,'' \emph{Report x-io and University of Bristol (UK)}, vol.~25, pp. 113--118, 2010.

\bibitem{Moduler}
S.~Lynen, M.~W. Achtelik, S.~Weiss, M.~Chli, and R.~Siegwart, ``A robust and modular multi-sensor fusion approach applied to mav navigation,'' in \emph{2013 IEEE/RSJ International Conference on Intelligent Robots and Systems}, pp. 3923--3929.

\bibitem{s111009182}
A.~M. Sabatini, ``Kalman-filter-based orientation determination using inertial/magnetic sensors: Observability analysis and performance evaluation,'' \emph{Sensors}, vol.~11, no.~10, pp. 9182--9206, 2011.

\bibitem{DBLP:conf/wafr/VarinK18}
P.~Varin and S.~Kuindersma, ``A constrained kalman filter for rigid body systems with frictional contact,'' in \emph{Algorithmic Foundations of Robotics XIII, Proceedings of the 13th Workshop on the Algorithmic Foundations of Robotics, {WAFR} 2018}, ser. Springer Proceedings in Advanced Robotics, vol.~14.\hskip 1em plus 0.5em minus 0.4em\relax Springer, 2018, pp. 474--490.

\bibitem{FRISON201580}
G.~Frison, M.~Vukov, N.~{Kjølstad Poulsen}, M.~Diehl, and J.~{Bagterp Jørgensen}, ``High-performance small-scale solvers for moving horizon estimation,'' \emph{IFAC-PapersOnLine}, vol.~48, no.~23, pp. 80--86, 5th IFAC Conference on Nonlinear Model Predictive Control NMPC 2015.

\bibitem{osqp}
B.~Stellato, G.~Banjac, P.~Goulart, A.~Bemporad, and S.~Boyd, ``{OSQP}: an operator splitting solver for quadratic programs,'' \emph{Mathematical Programming Computation}, vol.~12, no.~4, pp. 637--672, 2020.

\end{thebibliography}

\end{document}